\documentclass[letterpaper, 10 pt, conference]{ieeeconf}

\IEEEoverridecommandlockouts
\usepackage{amsmath}
\usepackage{amssymb}

\usepackage{amsthm}
\usepackage{graphicx}
\usepackage{balance}
\usepackage{multirow}
\usepackage{siunitx}
\usepackage{url}
\usepackage[hidelinks]{hyperref}
\hypersetup{
  pdftitle={SoftCorr: Differentiable Local Correlation for Single-Frame Targetless LiDAR-Camera Calibration},
  pdfauthor={Yuhyun Kim, Junseong Park, Dong-Geol Choi}
}

\usepackage{eso-pic}
\AddToShipoutPictureFG*{%
  \AtPageLowerLeft{%
    \put(\LenToUnit{0.5\paperwidth},\LenToUnit{0.42in}){%
      \makebox[0pt][c]{\parbox{\textwidth}{\centering\footnotesize
        This work has been submitted to the IEEE for possible publication.
        Copyright may be transferred without notice, after which this version may no longer be accessible.}}}}}

\newcommand{\ind}{\mathbf{1}}

\newcommand{\SEthree}{\mathrm{SE}(3)}
\newcommand{\sethree}{\mathfrak{se}(3)}
\newcommand{\neff}{n_{\mathrm{eff}}}
\newcommand{\Tinit}{T_{\mathrm{init}}}
\newcommand{\Sgeo}{\mathcal{S}_{\mathrm{geo}}}
\newcommand{\NID}{\overline{\mathrm{NID}}}

\title{\LARGE \bf
SoftCorr: Differentiable Local Correlation for \\
Single-Frame Targetless LiDAR--Camera Calibration
}

\author{Yuhyun Kim$^{1}$, Junseong Park$^{1}$, and Dong-Geol Choi$^{1,*}$%
\thanks{$^{1}$The authors are with the Department of Information and Communication Engineering,
Hanbat National University, 125 Dongseo-daero, Yuseong-gu, Daejeon 34158, Republic of Korea.
{\tt\small \{yuhyun.kim, junseong.park\}@edu.hanbat.ac.kr, dgchoi@hanbat.ac.kr}}%
\thanks{$^{*}$Corresponding author: Dong-Geol Choi.}%
}

\begin{document}

\maketitle
\thispagestyle{empty}
\pagestyle{empty}

\begin{abstract}
Targetless LiDAR--camera calibration from one image and one sweep needs no target, no scene
reconstruction, and no training on a dataset's recorded extrinsic. The strongest methods in this
setting correlate monocular depth against projected LiDAR range inside image patches. On that
fixed partition a point's whole contribution jumps whenever the pose carries it across a patch
edge, so the objective is rough, is searched rather than descended, and the search fails as the
initialisation worsens. We show that the
roughness is a property of the partition, not of the statistic. A local correlation depends on
its inputs only through six sufficient statistics per point, and SoftCorr accumulates those
continuously, by bilinear splatting and a separable kernel, which places an estimable cell at
every pixel and makes the objective differentiable almost everywhere. Calibration becomes
descent. With one set of constants fixed on KITTI and carried unchanged to four benchmarks,
SoftCorr converges on $365$ of $400$ single-frame KITTI calibrations against $238$ for the
closest released prior work re-run on identical frames, and on $310$ against $121$ from a
\SI{22.33}{\degree} start. With the depth network pinned, a controlled decomposition attributes
the largest share of the estimator's margin to the derivative itself: on the prior work's own
disjoint layout, descending our objective already beats searching it by more than the field
then adds. Code is available at \url{https://github.com/yuhyun00/SoftCorr}.
\end{abstract}


\section{Introduction}

Every LiDAR--camera fusion pipeline rests on the extrinsic between the two sensors, as does
any task that fits itself to the projected sweep at test time, such as test-time adaptation
for depth completion~\cite{lee2025zeroshotdc,seo2026lowrankdc}. That extrinsic
drifts, so it must be recoverable in the field, from whatever the two sensors happen to see. Target-based
procedures~\cite{zhang2004extrinsic} need a prepared scene, and learned
regressors~\cite{schneider2017regnet} are fitted to one dataset's recorded extrinsic. The targetless statistical-similarity line
avoids both. It began with mutual information between LiDAR reflectivity and image
intensity~\cite{pandey2015automatic}; its modern form adds a geometric signal, a monocular
network~\cite{yang2024depthanythingv2,wang2025moge2} whose dense depth prediction the
projected sweep must agree with. One image and one sweep suffice, with no scene model and no
labels.

\begin{figure}[!t]
\centering
\includegraphics[width=\linewidth]{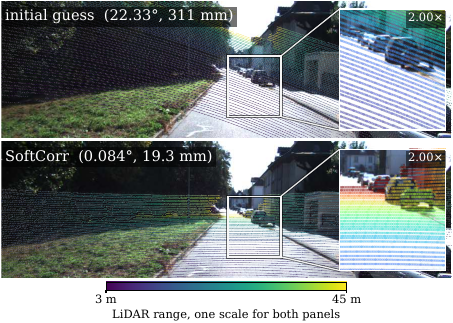}
\caption{A KITTI frame with the sweep projected and coloured by range, at the prior work's
rough-guess start (top, \SI{22.33}{\degree} composed) and at SoftCorr's single-frame estimate
from it (bottom). The \SI{22.33}{\degree} row converges on $38$ of $40$ frames of this sequence
(00), so the trial is typical rather than chosen.}
\label{fig:teaser}
\end{figure}

\textbf{Why these methods search.} A monocular network's output is not metric depth: its scale
and offset are unknown and vary across the image, so the correlation with projected range must
be local, computed inside patches that each carry their own affine nuisance. Locality is bought
with a partition, and the partition is what makes the objective rough: a return's whole
contribution moves to the next patch the instant the pose carries it across an edge, and
summed over tens of thousands of returns the score is a staircase in the pose
(Fig.~\ref{fig:landscape}). Without a usable derivative this family
searches, and a search has a capture range set by its budget, so it degrades sharply as the
initialisation worsens. The scene-model line~\cite{herau2024gscalib,zhang2025higscalib,kim2026bridging}
descends a differentiable rendering objective instead, but pays for the gradient with a
per-scene reconstruction over tens of frames.

\begin{figure*}[!t]
\centering
\includegraphics[width=\textwidth]{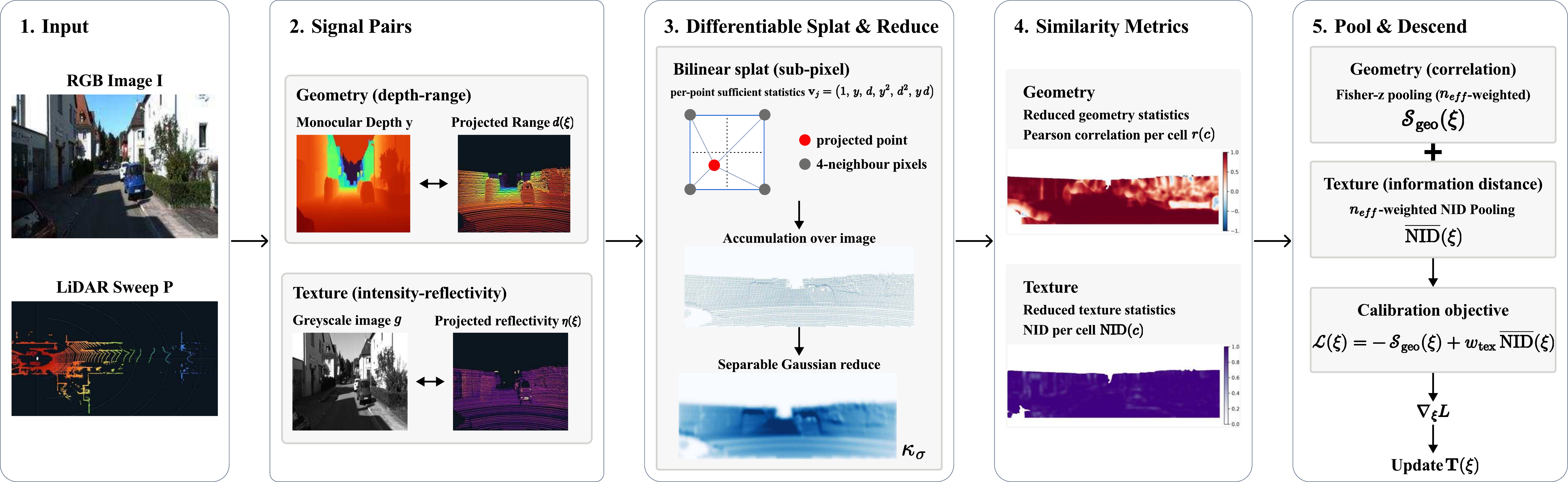}
\caption{SoftCorr. (1) One image and one sweep. (2) Monocular depth $y=Y(u)$, the network's
map sampled at the projected points, against projected range $d(\xi)$, and greyscale $g$
against projected reflectivity $\eta(\xi)$. (3) Per-point sufficient statistics $v_j$,
splatted bilinearly and reduced with a separable Gaussian $\kappa_\sigma$: a cell at every
pixel. (4) Per cell, Pearson's $r(c)$ and $\mathrm{NID}(c)$. (5) $\neff$-weighted pooling
gives $\Sgeo(\xi)$ and $\NID(\xi)$, which form $\mathcal{L}(\xi)$~\eqref{eq:loss}; its
gradient updates $T(\xi)$.}
\label{fig:pipeline}
\end{figure*}

\textbf{This paper.} Nothing in the statistic requires the partition. A local Pearson
correlation is a function of six moments per point, and those moments can be accumulated
continuously, each return spread over the pixels it falls between and pooled by a smooth
kernel. The result, which we call \emph{SoftCorr}, is a correlation field with an estimable
cell at every pixel, whose score is differentiable almost everywhere in the pose
(Fig.~\ref{fig:pipeline}). Calibration becomes a descent problem, from one image and one sweep,
with nothing reconstructed. Against the closest released prior work re-run on identical frames
from its own initialisation, SoftCorr converges on $365$ of $400$ single-frame KITTI
calibrations against $238$, and the margin widens as the start worsens.

\smallskip\noindent\textbf{Contributions.}
\begin{enumerate}
\setlength{\itemsep}{2pt}\setlength{\parskip}{0pt}\setlength{\topsep}{2pt}
\item \textbf{A differentiable local-correlation field.} The non-smoothness of the
patch-correlation objective comes from its partition. Splatting the sufficient statistics and
reducing them with a separable kernel removes it: an estimable cell at every pixel and an
objective differentiable almost everywhere, with no scene model and no training.
\item \textbf{Where the margin comes from.} With the depth network and the frames pinned,
the derivative is the largest single term of the estimator's margin over the prior work,
larger than the objective or the cell layout, and the lead survives swapping the network in
either direction; overlapping cells help the search at an easy initialisation and descent at
a hard one.
\end{enumerate}

\section{Related Work}
\label{sec:related}

\textbf{Targetless calibration signals.} Targetless toolboxes align image intensity to LiDAR
reflectivity~\cite{koide2023general}, or match cross-modal masks and semantic
regions~\cite{huang2024mias,huang2025edo} at the price of a segmentation model; edge-based
methods need no learned model but need a scene with edges~\cite{levinson2013automatic,yuan2021pixel}.

\textbf{Monocular depth as the bridge.} Correlating predicted depth against projected
range~\cite{borer2024chaos,zhang2025claim} turns a cross-modal problem into a same-modality
one. Borer et al.~\cite{borer2024chaos} treat the objective as derivative-free, the perspective
projection appearing to preclude an analytic gradient; Sec.~\ref{sec:field} shows the obstacle
is the binning, not the projection. CLAIM~\cite{zhang2025claim} is the closest prior work and
defines the recipe we start from: Pearson correlation over two \SI{40}{px} patch grids offset by
half a patch, between Depth Anything V2~\cite{yang2024depthanythingv2} inverse depth and the
LiDAR projection; a normalised information distance between greyscale and reflectivity; and
coarse-to-fine random search. We keep both signals, change only where the statistic is
evaluated, and re-run CLAIM under its own protocol, so a paired comparison isolates that change.
It is the only released method that reads the same signal (Sec.~\ref{sec:baseline} reports
what the others return on our frames).

\textbf{Scene-model calibration.} A second line reconstructs the scene and optimises a
differentiable-rendering proxy jointly with the
extrinsic~\cite{herau2024gscalib,zhang2025higscalib,zhou2025robust2dgs,kim2026bridging}.
The gradient is bought with that reconstruction, built from $30$ to $50$ frames per
estimate~\cite{herau2024gscalib,zhang2025higscalib,zhou2025robust2dgs}, where our unit is one
image and one sweep; matching either protocol would measure the protocol rather than the
estimator, so we rank neither line against the other. To our knowledge the similarity statistic itself has not
previously been made differentiable from a single frame with nothing reconstructed.

\textbf{Windowed correlation and degeneracy.} Convolved sufficient
statistics~\cite{cachier2000gaussian}
and overlapping NDT cells~\cite{biber2003ndt} are precedents, but in none of them does the pose
move a sparse point set and so change \emph{which} cells are estimable
(Sec.~\ref{sec:ablation}). That planar geometry leaves pose directions unconstrained is
classical~\cite{zhang2016degeneracy}; here it is absorbed by per-cell nuisance
parameters before any normal matrix exists (Sec.~\ref{sec:limits}).

\section{Method}

\subsection{Problem Setup and Locality}
\label{sec:locality}

Let $\{p_j\}\subset\mathbb{R}^3$ be one camera-facing LiDAR sweep (about \num{60000} returns on
KITTI), $I$ the hardware-triggered image, and $K$ its pinhole intrinsics with focal length $f$.
We seek $T\in\SEthree$ mapping LiDAR to camera, parameterised as a left perturbation
$T(\xi)=\exp(\xi^{\wedge})\,\Tinit$ with $\xi=(\rho,\phi)\in\sethree$. A frozen monocular
network supplies a dense map $Y$, and the pose reaches the data through
\begin{equation}
\label{eq:measure}
\begin{gathered}
z_j(\xi)=[T(\xi)p_j]_z,\qquad u_j(\xi)=\pi\big(K\,T(\xi)p_j\big),\\[2pt]
y_j(\xi)=Y\big(u_j(\xi)\big),\qquad d_j(\xi)=\psi\big(z_j(\xi)\big),
\end{gathered}
\end{equation}
where $Y$ is read by bilinear sampling and $\psi$ places LiDAR depth in the network's space
($\psi=\mathrm{id}$ for a metric checkpoint, $1/z$ for a relative one). $Y$ is rescaled to
about $\pm1$ by its 2/98 percentiles, which fixes the units of the variance floor in
Sec.~\ref{sec:field}.

\textbf{Why local.} A global correlation would suffice if $y=ad+b$ held with unknown $a,b$; it
does not, since monocular networks compress range non-uniformly and their error varies smoothly
with image position. The relation does hold inside a cell with its own $(a_c,b_c)$, so we
correlate per cell at the cost of a nuisance pair per cell~\cite{neyman1948consistent}.

\textbf{What locality costs.} A pose perturbation moves a cell's projected depth by
$\rho_z+z(\phi_xu^{(2)}-\phi_yu^{(1)})/f$, with $u^{(1)},u^{(2)}$ measured from the principal
point, so on a metric checkpoint a cell of width $\delta_c$ centred at $\bar u$ sees
\begin{equation}
\label{eq:transport}
\begin{split}
d_j&\;\mapsto\;\alpha_c\,d_j+\beta_c+d_j\,O(\|\phi\|\delta_c/f),\\[2pt]
\alpha_c&=1+(\phi_x\bar u^{(2)}-\phi_y\bar u^{(1)})/f,\qquad \beta_c=\rho_z.
\end{split}
\end{equation}
The leading part is exactly the affine family the nuisance pair absorbs. To that order the depth
channel carries pooling weight and no pose information; what carries the pose is $y$ resampled
at moved coordinates; depth still decides which cells are estimable at all, since a cell whose
returns share one depth has no correlation to measure (Sec.~\ref{sec:limits}).

\subsection{Lifting the Statistic off the Grid}
\label{sec:field}

A partition makes the cell both the unit of estimation and the unit of resolution, as in voxel
registration~\cite{koide2021vgicp}: a grid fine enough to resolve small displacements cannot
keep its cells estimable. SoftCorr never assigns a point to a cell. Every pixel anchors a cell,
built from the same six per-point quantities,
\begin{equation}
v_j=\big(1,\;y_j,\;d_j,\;y_j^2,\;d_j^2,\;y_jd_j\big),
\end{equation}
with $w_j\in\{0,1\}$ switching off points behind the near plane or outside the image
(Sec.~\ref{sec:terms}). Writing $\mathcal{B}[\cdot]$ for the bilinear splat, $\kappa_\sigma$
for separable Gaussian taps of width $\sigma$, and $\kappa_\sigma^{\odot2}$ for their
elementwise square,
\begin{equation}
\label{eq:reduce}
\begin{split}
M(c)=\big(\kappa_\sigma * \mathcal{B}[w\,v]\big)(c),\\[2pt]
Q(c)=\big(\kappa_\sigma^{\odot 2} * \mathcal{B}[w^2]\big)(c),
\end{split}
\end{equation}
evaluated at every pixel $c$. Cells overlap and each point contributes to many. Splat and
reduction both zero-pad, so the effective sample size falls off at the border instead of being
renormalised. The pose enters once, at $T(\xi)$, so the loop closes through a gradient.

The six channels of $M$ are the moments a weighted Pearson correlation needs, so $r(c)$ is
assembled in closed form at every $c$ (means from $M_y$, $M_d$; variances and covariance from
$M_{y^2}$, $M_{d^2}$, $M_{yd}$, each variance a positive part above a floor of $10^{-6}$). Write
$M_0(c)=\sum_j w_j\kappa_j(c)$ for the splatted weight, $\kappa_j(c)$ being the tap weight point
$j$ deposits at $c$.

\begin{figure*}[!t]
\centering
\includegraphics[width=\textwidth]{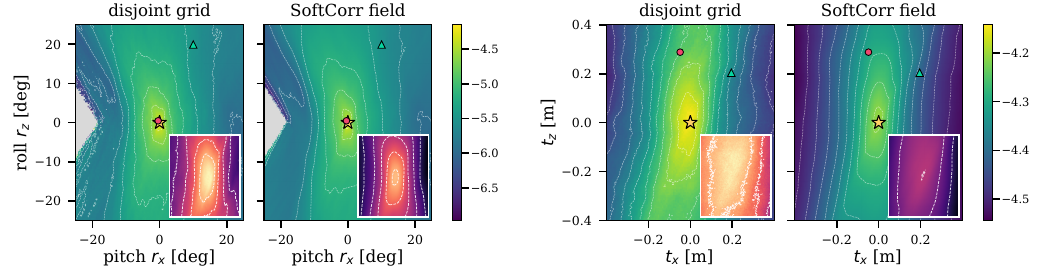}
\caption{The objective on the frame of Fig.~\ref{fig:teaser} over two pose coordinates: the
disjoint box grid (left of each pair) and the SoftCorr field (right) at one pinned $\sigma$.
Star: reference pose; triangle: the prior work's \SI{22.33}{\degree} rough-guess start; circle:
the nominal start. Insets: \SI{\pm1}{\degree} and \SI{\pm0.05}{\metre} about the reference,
with contours. Grey: poses at which too few cells are estimable for a score to exist.}
\label{fig:landscape}
\end{figure*}

\begin{figure}[!tb]
\centering
\includegraphics[width=\linewidth]{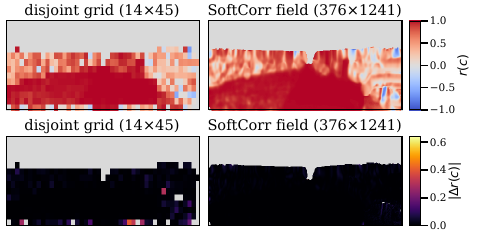}
\caption{Per-cell correlation on the frame of Fig.~\ref{fig:teaser} at the same $\sigma$,
disjoint grid (left) against field (right). Top: $r(c)$ at the reference pose, cells with
$\neff<n_{\min}$ in grey. Bottom: $|\Delta r(c)|$ under a rotation worth \SI{0.3}{px} in the
image.}
\label{fig:rmap}
\end{figure}

\textbf{What the splat changes.} A cell's contents now change smoothly under a displacement
smaller than one cell. Fig.~\ref{fig:landscape} draws the objective on one frame over pitch
and roll (\SI{\pm25}{\degree}) and over $t_x$ and $t_z$ (\SI{\pm0.4}{\metre}), the disjoint
box grid against the Gaussian field we ship at one pinned $\sigma=\SI{8}{px}$ (cell width
\SI{28}{px}). At the scale of the capture range the two share one basin; inside a cell the
grid is a staircase and the field is smooth, the detrended residual of the inset being
$14\times$ larger on the grid in rotation and $20\times$ in translation. Isolating the stride
alone, with the box kernel on both sides, a one-dimensional \SI{\pm1}{px} slice reads
$18\times$ at the median over five sequences, and $27\times$ against the Gaussian field.
Fig.~\ref{fig:rmap} shows the same frame at the cell level, $14\times45$ cells on the grid
against one per pixel on the field, where a rotation worth \SI{0.3}{px} (\SI{0.024}{\degree}) makes single grid cells jump
by $0.2$--$0.6$ as a return crosses an edge while every field cell that saw the return moves a
little (mean $|\Delta r|$ $2.7\times$ smaller). Over $24$ poses
the analytic gradient agrees with a central difference of the same objective at a median
cosine of $0.97$ on the field against $0.78$ on the grid: the derivative the splat provides is
faithful to the objective.

\textbf{Non-smooth points.} Every operation is smooth in $\xi$ except three benign kinds:
bilinear interpolations ($C^0$, derivative jumps averaged over $2\times10^4$ returns), the
clamp and variance positive parts ($C^0$), and the $n_{\min}$ and visibility indicators,
discontinuous with identically zero gradient.

\subsection{Effective Sample Size and Pooling}
\label{sec:pooling}

Cells differ by orders of magnitude in how many returns support them. The relevant count is
Kish's effective sample size~\cite{kish1965survey},
\begin{equation}
\neff(c)=\frac{M_0(c)^2}{Q(c)},
\label{eq:neff}
\end{equation}
which is why $Q$ carries the elementwise square of the taps in~\eqref{eq:reduce}: $\neff$
equals the point count under equal weights and collapses to $1$ when one point dominates. $Q$
squares the taps but not the splat weights, so it upper-bounds the exact Kish denominator
$\sum_j(w_j\kappa_j(c))^2$ and $\neff$ is a conservative underestimate, tight for
$\sigma\gg\SI{1}{px}$. Since
$\operatorname{Var}[\operatorname{artanh}r]\approx 1/(\neff-3)$ independently of the true
correlation~\cite{fisher1921probable}, we pool by it:
\begin{equation}
\begin{split}
\Sgeo(\xi)=
\frac{\sum_c \omega(c)\,\zeta\big(r(c)\big)}{\sum_c \omega(c)},\\[2pt]
\zeta(r)=\operatorname{artanh}\big(\mathrm{clip}\,r\big),
\label{eq:pool}
\end{split}
\end{equation}
\begin{equation}
\label{eq:omega}
\omega(c)=\big(\neff(c)-3\big)_+\,\ind\big[\neff(c)\ge n_{\min}\big].
\end{equation}
The clip is a hard clamp at $\pm0.9995$, forced because $r$ leaves $[-1,1]$
when a cell's variance sits on its floor. We call $\Sgeo$ the \emph{geometric term}.

\subsection{Cell Scale}
\label{sec:bandwidth}

Return density varies by an order of magnitude across the image, so we fix the neighbour
count rather than the width~\cite{fan1996local}: $\sigma$ is bisected over \SIrange{2}{64}{px}
until the frame-averaged median $\neff$ over cells with $\neff>1$ reaches $48$, once per
calibration at its own initialisation. The solution $\sigma^{\star}$ is a median \SI{7.7}{px}
over the shipped runs (\SIrange{5.1}{13.2}{px} across all configurations). The rule sets the
scale from the data rather than from the dataset.

\subsection{Texture, Visibility, and Optimisation}
\label{sec:terms}

\textbf{Texture.} Geometry is blind to anything flat, so we add the photometric channel
of~\cite{zhang2025claim}: greyscale $g$ and reflectivity $\eta$ are rank-equalised and
soft-binned into $N_b=6$ bins with linear B-splines, the joint channels are splatted and
reduced as above, and per cell we take the normalised information distance
$1-\mathrm{MI}(g,\eta)/H(g,\eta)$~\cite{studholme1999overlap}, pooled by the same $\omega(c)$
and written $\NID$. With $w_{\text{tex}}=6$,
\begin{equation}
\mathcal{L}(\xi)=-\,\Sgeo(\xi)+w_{\text{tex}}\,\NID(\xi).
\label{eq:loss}
\end{equation}

\textbf{Visibility.} A hard mask keeps points at least $m=4$ pixels inside the image border.

\textbf{Optimisation.} We minimise~\eqref{eq:loss} with AdamW ($\mathrm{lr}=5\times10^{-3}$)
over a coarse-to-fine ladder $\{8,6,4,3,2,1.5,1\}\times\sigma^{\star}$, $90$ steps per rung
($630$ in total), optimiser state re-initialised at each rung. Descent returns its final
iterate; nothing is selected on the score.

\textbf{Constants.} Every constant above ($n_{\min}$, the clip, the variance floor, the
$\neff$ target, $N_b$, $w_{\text{tex}}$, $m$, the ladder and the step size) was fixed once on
KITTI odometry sequences 00 and 01 and is used unchanged on every benchmark and rig in this
paper.

\section{Experiments}
\label{sec:exp}

\subsection{Setup}
\label{sec:protocol}

\textbf{Data.} Four benchmarks, all defined by prior work. KITTI odometry~\cite{geiger2012kitti}
00--09 carries the paired comparison against released code; the ten KITTI raw drives and first
ten Waymo scenarios~\cite{sun2020waymo} of~\cite{zhang2025claim} carry its rough-guess
benchmark; MIAS-LCEC~\cite{huang2024mias} is a second rig, a Livox Mid-70 (TF70) and a Mid-360
(TF360), both non-repetitive scanners, with a $1200\times800$ camera with real distortion.
Every calibration consumes \emph{one} image and \emph{one} sweep unless a row says otherwise.

\textbf{Initialisation and frames.} The nominal axis permutation with zero translation, shared
by~\cite{koide2023general,huang2024mias,huang2025edo,zhang2025claim}; KITTI odometry's three
recorded extrinsics make this three starting errors (\SIlist{0.62;0.85;0.60}{\degree},
\SIlist{298;286;341}{\milli\metre}), and every paired test is formed within a sequence. Trial
$k$ draws one frame uniformly from the sequence under a fixed seed, giving $400$ calibrations
that every arm sees. The released baseline runs at its
published defaults and, as a control, at the ranges its documentation recommends; our
constants are the single set of Sec.~\ref{sec:terms}, chosen on sequences 00 and 01. Every
count is over all ten sequences except Table~\ref{tab:attrib}(b), Fig.~\ref{fig:cell} and the
search control of Sec.~\ref{sec:ablation}, which use the eight held out from the step-size
sweeps ($320$ trials).

\textbf{Metrics.} Errors are read off $\Delta T=\hat T\,T_{\mathrm{ref}}^{-1}$, so the translation
error is $\|\Delta T_t\|$; both methods are scored identically. A trial is \emph{converged}
below \SI{1.0}{\degree} and \SI{100}{\milli\metre} (the prior work's own rotation threshold and
half its translation one). The body reports two
statistics: the \emph{convergence count} of $400$, compared by exact McNemar on the discordant
pairs, and the \emph{paired accuracy}, the median per-frame difference in pose error on the
frames both arms converge, compared by paired Wilcoxon. Both pool correlated frames as though
independent and are uncorrected for multiplicity. The splat uses GPU atomics and is not
bit-reproducible: five runs of the shipped configuration read $364$--$367$ of $400$ and the baseline at three seeds reads
$222$--$238$, so contrasts of that size are not read as effects.

\textbf{Cost.} One NVIDIA RTX A6000, MoGe-2~\cite{wang2025moge2} as the depth network. A
calibration takes \SI{31.5}{\second} plus \SI{21}{\milli\second} of cached monocular inference,
against the baseline's \SI{22.4}{\second}; at the $448$ steps that match its wall clock we
converge $363/400$, so the headline is not bought with the longer schedule.

\begin{table*}[!t]
\centering
\caption{Paired comparison on identical frames: ten KITTI odometry sequences, $40$ trials each;
$e_r$ in \si{\degree} (leading zero dropped) and $e_t$ in \si{\milli\metre}, medians over the
trials each method converges on; last column, trials inside \SI{1.0}{\degree} and
\SI{100}{\milli\metre} (bold: more). SF: one frame; MF: four from the nominal initialisation, both arms on the same drawn frames,
the baseline in its own four-frame mode. CLAIM$^{\dagger}$: the released code re-run on our frames.}
\label{tab:main}
\scriptsize
\setlength{\tabcolsep}{1.0pt}
\begin{tabular*}{\linewidth}{@{\extracolsep{\fill}}@{}cc|cc|cc|cc|cc|cc|cc|cc|cc|cc|cc|c@{}}
\hline
\multicolumn{2}{c|}{\multirow{2}{*}{Method}} & \multicolumn{2}{c|}{00} & \multicolumn{2}{c|}{01} & \multicolumn{2}{c|}{02} & \multicolumn{2}{c|}{03} & \multicolumn{2}{c|}{04} & \multicolumn{2}{c|}{05} & \multicolumn{2}{c|}{06} & \multicolumn{2}{c|}{07} & \multicolumn{2}{c|}{08} & \multicolumn{2}{c|}{09} & \multirow{2}{*}{conv.} \\ \cline{3-22}
\multicolumn{2}{c|}{} & $e_r$ & $e_t$ & $e_r$ & $e_t$ & $e_r$ & $e_t$ & $e_r$ & $e_t$ & $e_r$ & $e_t$ & $e_r$ & $e_t$ & $e_r$ & $e_t$ & $e_r$ & $e_t$ & $e_r$ & $e_t$ & $e_r$ & $e_t$ & \\ \hline
\multirow{3}{*}{SF} & \multicolumn{22}{@{}l}{\emph{nominal initialisation, \SIrange{0.60}{0.85}{\degree}}} \\ \cline{2-23}
& CLAIM$^{\dagger}$ & .267 & 55.2 & .151 & 83.2 & .251 & 63.0 & .293 & 75.3 & .255 & 65.2 & .289 & 55.3 & .223 & 71.3 & .294 & 58.2 & .339 & 64.6 & .305 & 57.2 & 238/400 \\
& Ours & .088 & 20.2 & .120 & 72.7 & .096 & 28.7 & .169 & 37.5 & .140 & 37.0 & .179 & 32.0 & .142 & 29.0 & .183 & 38.4 & .200 & 31.5 & .172 & 30.8 & \textbf{365}/400 \\ \hline
\multirow{3}{*}{SF} & \multicolumn{22}{@{}l}{\emph{their \SI{10}{\degree}/\SI{0.2}{\metre}-an-axis recipe, \SI{22.33}{\degree} composed}} \\ \cline{2-23}
& CLAIM$^{\dagger}$ & .233 & 49.1 & .163 & 73.2 & .244 & 48.9 & .094 & 55.0 & .195 & 65.3 & .274 & 62.7 & .326 & 62.0 & .364 & 66.9 & .302 & 54.7 & .289 & 54.9 & 121/400 \\
& Ours & .092 & 19.1 & .123 & 55.7 & .096 & 29.1 & .173 & 36.5 & .137 & 37.6 & .185 & 34.6 & .142 & 29.0 & .192 & 39.4 & .206 & 31.6 & .176 & 32.6 & \textbf{310}/400 \\ \hline
\multirow{2}{*}{MF} & CLAIM-4F$^{\dagger}$ & .171 & 34.3 & .122 & 61.2 & .154 & 42.4 & .198 & 48.0 & .159 & 37.2 & .235 & 41.4 & .280 & 53.9 & .243 & 46.7 & .267 & 50.6 & .181 & 38.5 & 370/400 \\
& SoftCorr-4F & .060 & 14.1 & .095 & 57.9 & .054 & 27.3 & .153 & 36.0 & .124 & 22.8 & .161 & 23.3 & .145 & 22.0 & .161 & 27.6 & .188 & 27.8 & .160 & 27.3 & \textbf{400}/400 \\
\hline
\end{tabular*}
\end{table*}

\subsection{Paired Comparison with the Released Baseline}
\label{sec:baseline}

In Table~\ref{tab:main} the released prior work runs on our frames from our initialisation
and is scored by our code, so its rows differ only in the estimator and in the depth backbone
each ships with, which Sec.~\ref{sec:attrib} separates. Our re-run reproduces the baseline's
published sequence-00 figure under its own ungated mean (\SI{0.365}{\degree} against
\SI{0.366}{\degree}).

\textbf{The other released baselines,} run on the same frames at their published defaults,
do not reach the gate (Koide et al.~\cite{koide2023general} $0/400$, and $4/400$ and $2/400$
with their \SI{0.2}{\metre} trust region widened to \SI{0.5}{\metre} and \SI{1}{\metre};
HKU-Mars~\cite{yuan2021pixel} $0$ of the $389$ frames it returns), so CLAIM is the only arm a paired comparison can be run against.

\textbf{Reachability.} SoftCorr converges on $365$ of $400$ calibrations against $238$. The
discordant pair is $138$ to $11$ ($p=4.2\times10^{-29}$), and the lead holds on $10/10$
sequences. The margin does not depend on the gate: it is $55$ frames at the baseline's own
\SI{200}{\milli\metre} translation threshold and grows as the gate tightens, to $288$ against
$67$ at \SI{0.5}{\degree}/\SI{50}{\milli\metre}. Nor is it an artefact of how the baseline was run:
at the ranges its documentation recommends it converges $243/400$, and with its coarse
translation search widened past our perturbation $234/400$ ($p = 0.78$).

\textbf{Accuracy.} Where both arms converge, ours is the more accurate by a paired
\SI{+0.125}{\degree} and \SI{+28.2}{\milli\metre} on $227$ frames, signed our way on all ten
sequences; restricting the medians of Table~\ref{tab:main} to those frames changes no
column's sign. Both arms are scored against the same recorded extrinsic, so its day-to-day drift
(\SI{0.101}{\degree}, \SI{21.0}{\milli\metre}) cancels in the paired difference.

\textbf{Capture range.} The prior work's protocol adds $(\epsilon,\epsilon,\epsilon)$ degrees to
the recorded Euler angles and $(t,t,t)$ metres to the translation. We run five such rungs,
$(\epsilon,t)=(1,0.05)$, $(2,0.1)$, $(5,0.2)$, $(10,0.2)$, $(15,0.3)$, labelled by composed
rotation; $(10,10,10)$ composes to \SI{22.33}{\degree} and \SI{311}{\milli\metre} in our chart
against \SI{17.75}{\degree} in the prior work's yaw--pitch--roll chart, and we run the harder,
both arms from the identical pose. The shipped configuration converges on $368$, $368$, $352$,
$310$ and $180$ of $400$ across the rungs; at \SI{22.33}{\degree} the baseline reaches
$121/400$ against our $310$ (Table~\ref{tab:main}), discordant $203$ to $14$ on $10/10$
sequences.

\begin{table}[!t]
\centering
\caption{Where the margin comes from: converged calibrations at the nominal initialisation.
(a) Both estimators on both depth backbones, all ten sequences ($400$ trials). (b) The
estimator axis with the backbone pinned to MoGe-2, on the eight sequences held out from the
step-size sweeps ($320$ trials), from the released baseline to the method we ship. Bold: best.}
\label{tab:attrib}
\footnotesize
\setlength{\tabcolsep}{3pt}
\begin{tabular*}{\linewidth}{@{\extracolsep{\fill}}@{}lcc@{}}
\hline
\multicolumn{3}{@{}l}{\emph{(a) backbone axis, of $400$}} \\
estimator & DA-V2~\cite{yang2024depthanythingv2} & MoGe-2~\cite{wang2025moge2} \\
\hline
CLAIM (released code) & 238 & 294 \\
SoftCorr (ours) & 321 & \textbf{365} \\
discordant, ours/CLAIM & $111$/$28$ & $84$/$13$ \\
\hline
\multicolumn{3}{@{}l}{\emph{(b) estimator axis, MoGe-2 for every arm, of $320$}} \\
\multicolumn{2}{@{}l}{arm} & conv. \\
\hline
\multicolumn{2}{@{}l}{CLAIM: its objective, its patch grid, its search} & 240 \\
\multicolumn{2}{@{}l}{our objective on a disjoint grid, equal-budget search} & 256 \\
\multicolumn{2}{@{}l}{same objective and grid, \emph{descent}} & 295 \\
\multicolumn{2}{@{}l}{the field we ship, descent} & \textbf{301} \\
\hline
\end{tabular*}
\end{table}

\subsection{Where the Margin Comes From}
\label{sec:attrib}

Table~\ref{tab:attrib} tests the paper's claim by moving from the released prior work to the
method we ship one axis at a time: the depth backbone, then the objective, the optimiser and
the layout.

\textbf{The backbone is not the margin.} Our headline runs on MoGe-2 where the baseline ships
with Depth Anything V2, so Table~\ref{tab:attrib}(a) runs both estimators on both networks. The
backbone is worth $56$ frames to the baseline and $44$ to ours (interaction $-12$,
sequence-level interval $[-26,0]$); both within-column contrasts are signed our way on $10/10$
sequences ($p=6.9\times10^{-14}$ for the MoGe-2 pair). Of the $127$-frame headline margin, the network swap is worth
$56$ frames and is not ours; the rest is the estimator, which the next paragraph decomposes. All four backbones we ran beat the baseline's shipped configuration ($365$, $332$, $332$ and
$321$ for MoGe-2, UniDepth V2~\cite{piccinelli2025unidepthv2}, Depth Anything
3~\cite{lin2025depthanything3} and Depth Anything V2, against its $238$).

\textbf{The derivative is the largest term.} Table~\ref{tab:attrib}(b) pins the network and
walks from the baseline to our method in three steps: from the baseline's objective, patch
size and search to our objective on a disjoint grid at our cell width under an equal-budget
search (the CMA-ES arm of Sec.~\ref{sec:anchor}); from that search to descent; from the disjoint grid to the field. It reads baseline
$240$ of $320$; our objective, searched, $256$, descending it $295$; the field we ship $301$.
Read as differences, of those $61$ frames $16$ are the objective and the search we give it,
$39$ the derivative the splat makes available, and $6$ the stride, kernel and bandwidth
together, too small to read as an effect. Sequence 04
carries much of it: the search loses $15$ frames there against the baseline and descent
recovers them ($33$, $18$, $33$, $39$ down the rows). So about a quarter is bought before this paper's machinery is used, and we do
not claim it; the one change a disjoint grid cannot make, following the gradient, accounts for
most of the rest.

\subsection{What the Cell Layout Is Worth}
\label{sec:anchor}

\begin{figure}[!tb]
\centering
\includegraphics[width=\linewidth]{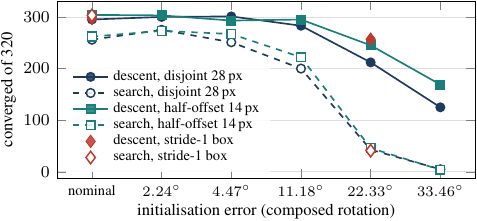}
\caption{Converged calibrations of $320$ on sequences 02--09 under descent (solid) and the
search (dashed), on disjoint \SI{28}{px} cells and half-offset \SI{14}{px} cells, box window
at one pinned $\sigma$ (cell width $\sqrt{4\pi}\,\sigma=\SI{28}{px}$). Diamonds: a box cell
at every pixel at the same $\sigma$, at two rungs.}
\label{fig:cell}
\end{figure}

Sec.~\ref{sec:field} makes two changes at once relative to a grid of patches: cells stop being
disjoint, and the window stops being a box. Fig.~\ref{fig:cell} holds one fixed while the
other moves, sweeping the stride of a box window at a pinned bandwidth so every arm is matched
on effective sample size, and runs each layout under both a search and descent. The search is
$(\mu/\mu_w,\lambda)$-CMA-ES~\cite{hansen2001cma} on the same objective at population $24$, at
$1344$ evaluations against descent's $630$ steps, which is equal arithmetic since a gradient
step costs $2.1$ forward evaluations. Every point has its step size swept at its own
layout and rung on sequences 00 and 01, each optimiser at its better schedule; the figure
reports 02--09, held out from those sweeps and from Sec.~\ref{sec:terms},
though not from the kernel and bandwidth rows of Table~\ref{tab:ablation}.
The stride-1 search at \SI{22.33}{\degree} keeps rising below its sweep's endpoint,
so its $41$ is a lower bound.

\textbf{Descent leads at every rung} of both line layouts ($95\,\%$ paired interval excludes
zero), by $25$ to $169$ of $320$ on the disjoint grid and $26$ to $199$ at one offset; the gap
peaks at \SI{22.33}{\degree}, past which the search is dead and descent is itself collapsing.
\textbf{What overlap buys depends on the pose.} Going from disjoint cells to a cell at every
pixel (diamonds) reverses between the two rungs. At the nominal pose it is worth $+47$ to the
search ($8/8$ sequences) and $+2$ to descent; at \SI{22.33}{\degree} it is worth $+45$ to
descent ($8/8$) and $-2$ to the search. At an easy start overlap helps only the optimiser that has no gradient; at
a hard one it helps only the one that does, whose basin the overlap widens. Consistently with
Table~\ref{tab:attrib}(b), descent at the nominal pose reads $295$ on the disjoint grid, $297$
at stride one, $304$ at the half offset and $301$ for the shipped field: there the layout moves
the count by less than ten frames of $320$ and the derivative by far more.

\subsection{Ablations and Sensitivity to the Constants}
\label{sec:ablation}

\begin{table}[!tb]
\centering
\caption{Top: single-component ablations at the nominal initialisation against the full
method at $365/400$ on identical frames; disc.: frames converged only by the full method /
only by the ablation; $\Delta$px: median paired reprojection difference;
bold: $p<0.05$ uncorrected. Bottom: each swept constant of Sec.~\ref{sec:terms} varied alone,
one run per value; spread: max$-$min over all runs, including replicates.}
\label{tab:ablation}
\scriptsize
\setlength{\tabcolsep}{2pt}
\begin{tabular*}{\linewidth}{@{\extracolsep{\fill}}@{}lccrr@{}}
\hline
\multicolumn{5}{@{}l}{\emph{ablation}} \\
change & conv. & disc. & $p$ & $\Delta$px \\
\hline
texture term off & 352/400 & 18/5 & \textbf{0.011} & $+0.049$ \\
geom.\ splat coords.\ detached & 294/400 & 73/2 & $\mathbf{1.4\!\times\!10^{-19}}$ & $+0.283$ \\
geom.\ pooling weight detached & 341/400 & 28/4 & $\mathbf{1.9\!\times\!10^{-5}}$ & $+0.155$ \\
bandwidth ladder removed & 364/400 & 11/10 & 1.00 & $+0.003$ \\
\hline
\multicolumn{5}{@{}l}{\emph{sensitivity (conv.\ of $400$)}} \\
constant & values & \multicolumn{2}{c}{conv.\ at each value} & spread \\
\hline
$\neff$ target $\neff^\star$ & $24/32/48/64$ & \multicolumn{2}{c}{$368/369/365/362$} & $7$ \\
kernel (pinned $\sigma$) & box / Gaussian & \multicolumn{2}{c}{$369/363$} & $6$ \\
bandwidth & rule / pinned $\sigma$ & \multicolumn{2}{c}{$365/363$} & $4$ \\
bins $N_b$ & $4/6$ & \multicolumn{2}{c}{$363/365$} & $4$ \\
\hline
\end{tabular*}
\end{table}

Table~\ref{tab:ablation} removes one component at a time (top) and varies each constant
(bottom). Two components matter. \textbf{Splat coordinates:} detaching them leaves the field
unchanged and removes the one derivative path a disjoint grid does not have; it costs
\emph{descent} $71$ frames while saying nothing about the field, which an equal-budget search
reaches with no derivative ($299$ against descent's $301$ of $320$). \textbf{Pooling weight:} the
extrinsic moves a sparse point set, so $\neff$ and the weight it sets are functions of $\xi$,
a derivative path with no counterpart in registration; it is a median \SI{20}{\percent} of the
gradient norm and acts as a shrinkage. The texture term is worth $13$ frames. The ladder is not required (Table~\ref{tab:ablation}; \SI{15.3}{\second} against
\SI{31.5}{\second} measured beside it without it);
we keep it so that descent and the search arm run one schedule. \textbf{Sensitivity.} Five runs of the identical shipped configuration read $364$--$367$,
and no swept constant moves convergence by more than $7$ frames, about twice that spread and
half the texture term; $n_{\min}$, the clip, the variance floor, the percentile scaling and the
steps per rung were not swept. The kernel row compares a box cell at every pixel with the
Gaussian at one pinned $\sigma$, so only the taps differ; it is the sweep the kernel was chosen
on, so it is a selection statistic, not a held-out one. The box reads $6$ frames higher at the
nominal pose and lower at the farthest rung ($26$ against $33$ of $80$ on sequences 00 and
01 at \SI{33.46}{\degree});
we ship the Gaussian for its smoothness (Sec.~\ref{sec:field}). The method is
not sitting on a tuned point.

\subsection{Transfer, and Where It Fails}
\label{sec:transfer}

\begin{figure*}[!t]
\centering
\includegraphics[width=\textwidth]{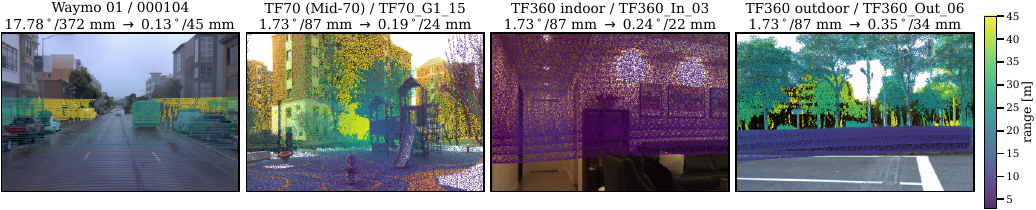}
\caption{The same constants on rigs the method was never tuned on: a Waymo scenario, a Livox
Mid-70 scene (MIAS-LCEC TF70) and two Mid-360 scenes (TF360, indoor in low light, and outdoor),
the sweep projected at SoftCorr's estimate and coloured by range; each panel states its own
initialisation and final error.}
\label{fig:transfer}
\end{figure*}

The constants of Sec.~\ref{sec:terms} were chosen on two KITTI sequences. This section runs
them unchanged on three benchmarks of the prior work and on a second rig, and reports the one
place they fail. The scenes of Fig.~\ref{fig:transfer} are the converged ones closest to each
benchmark's median rotation error, not the best, at the poses the reported runs produced.

\begin{table}[!tb]
\centering
\caption{The prior work's KITTI odometry benchmark, averaged over sequences 00--09: $e_r$ in
\si{\degree}, $e_t^-$ in \si{\metre}, ungated means with translation in the LiDAR frame (their
metric). Rows other than ours are from~\cite{zhang2025claim}; CLAIM-4F and CLAIM-4F$^{*}$ are
its two four-frame variants as it reports them. Bold: best in block; ``led'': sequences on
which the row is best in its block.}
\label{tab:proto}
\footnotesize
\setlength{\tabcolsep}{3pt}
\begin{tabular*}{\linewidth}{@{\extracolsep{\fill}}@{}clccc@{}}
\hline
& Method & $\bar e_r$ & $\bar e_t^-$ & led \\
\hline
\multirow{7}{*}{SF} & CRLF~\cite{ma2021crlf} & 0.642 & 6.554 & 0/10 \\
& UMich~\cite{pandey2015automatic} & 3.258 & 0.343 & 0/10 \\
& HKU-Mars~\cite{yuan2021pixel} & 25.417 & 6.079 & 0/10 \\
& DVL~\cite{koide2023general} & 121.33 & 4.487 & 0/10 \\
& MIAS-LCEC~\cite{huang2024mias} & 3.423 & 0.623 & 0/10 \\
& CLAIM~\cite{zhang2025claim} & 0.421 & 0.107 & 0/10 \\
& \textbf{SoftCorr (Ours)} & \textbf{0.200} & \textbf{0.053} & 10/10 \\
\hline
\multirow{4}{*}{MF} & EdO-LCEC~\cite{huang2025edo} & 0.666 & 0.169 & 0/10 \\
& CLAIM-4F~\cite{zhang2025claim} & 0.265 & 0.062 & 0/10 \\
& CLAIM-4F$^{*}$~\cite{zhang2025claim} & 0.226 & 0.051 & 0/10 \\
& \textbf{SoftCorr-4F (Ours)} & \textbf{0.131} & \textbf{0.029} & 10/10 \\
\hline
\end{tabular*}
\end{table}

\textbf{Published benchmarks.} Under each benchmark's own metric (ungated means, translation
in the LiDAR frame) and beside the numbers its authors published~\cite{zhang2025claim}: on
KITTI odometry (Table~\ref{tab:proto}) SoftCorr leads
every single-frame entry on all ten sequences, and four frames give the lowest error of any
entry including the multi-frame ones. On MIAS-LCEC-TF360 it leads the prior work's rows indoors, including its four-frame one
(\SI{0.364}{\degree}/\SI{39}{\milli\metre} against its \SI{0.457}{\degree}/\SI{46}{\milli\metre}),
and outdoors leads in translation but trails the multi-frame EdO-LCEC~\cite{huang2025edo} in
rotation (\SI{0.378}{\degree} against \SI{0.349}{\degree}).

\textbf{A second rig.} On all $71$ scenes of MIAS-LCEC, neither of whose scanners shares
sampling geometry with KITTI, $56$ converge from
\SI{1.73}{\degree}/\SI{87}{\milli\metre} and $39$ from \SI{8.66}{\degree}/\SI{261}{\milli\metre},
at medians of \SIlist{0.198;0.213}{\degree}. The \SI{87}{\milli\metre} start is already inside
the gate, so this rig tests rotation ($69$ scenes clear the rotation half, $56$ the translation
half). The bandwidth rule is not what transfers: KITTI's pixel ladder, run here unchanged, converges
the same $56$ scenes as the adaptive rule ($2$ discordant each way, $p=1$).

\begin{figure}[!tb]
\centering
\includegraphics[width=\linewidth]{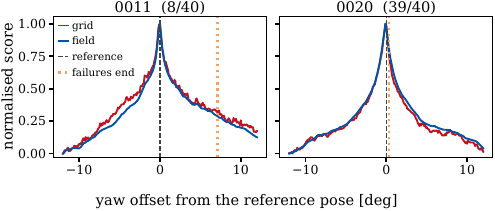}
\caption{The failure mode: the objective along the yaw axis through the reference pose on
one frame of KITTI raw drive 0011 (converges $8/40$) and one of drive 0020 ($39/40$), grid and
field at the pinned $\sigma$ of Fig.~\ref{fig:landscape}, each curve normalised to $[0,1]$.
Dashed: the reference; dotted: where that drive's failed trials end (on 0020 the two nearly
coincide).}
\label{fig:failure}
\end{figure}

\textbf{Where it fails.} On the prior work's rough-guess benchmark the same recipe is accurate
where it lands and narrow in where it lands. On the ten KITTI raw drives, over the $231$ of
$400$ trials that converge, the converged mean is below the prior work's own rows, single- and
four-frame, on all six components (a converged mean, which its ungated rows, stating no
convergence count, cannot be matched to like for like). The $169$ that fail sit at
\SI{6.48}{\degree} of yaw in the prior work's chart, near the \SI{10}{\degree} they started
at, and the drives differ widely: 0020 converges $39/40$ and 0011 $8/40$. Laddering two of
the drives locates it as a cliff, $75$, $75$ and $70$ of $80$ at \SIlist{1.73;3.47;8.75}{\degree}
falling to $27$ at \SI{17.74}{\degree}, the prior work's $(10,10,10)$ recipe in its chart, at
which the estimator reaches $310/400$ on odometry; neither our handling of the raw drives ($9$ of the
$10$ drives converge from the nominal guess) nor a harsher start explains it. What separates
the drives is how far away the scene is: a drive's median range and its converged count have
a Pearson correlation of $-0.911$ over the ten drives ($p=2\times10^{-4}$). The failing drives
are not the ones whose monocular depth is worst; depth agrees with true range
\emph{better} on them (median per-cell $r$ of $0.98$ on 0011 against $0.74$ on 0020). What
tracks convergence is the number of cells the pooling can use ($r=+0.745$, $p=0.013$), not
the number of returns ($r=+0.302$), and the share of pooling weight in planar cells is no
larger here than on odometry (\SI{43.5}{\percent} against \SI{41.5}{\percent}). This is
what~\eqref{eq:transport} predicts: a scene whose depth is near-affine in the image scores
well in every cell and carries little pose signal, so the basin is wide-shouldered rather than
absent. In Fig.~\ref{fig:failure} the reference is the global maximum on both drives and both
arms, and 0011's failed trials do not sit at a secondary maximum; its basin has a wider
shoulder and a rougher profile (detrended RMS $0.018$ against $0.015$ on 0020, grid). On Waymo
one frame converges $151$ of $400$ trials and four frames $257$, the converged mean below the
prior work's single-frame row on five of six components; four frames buy most of the range
back on the raw drives too ($341$ against $231$). These scenes cost capture range, not
accuracy.

\section{Limitations}
\label{sec:limits}

Capture range is finite ($180/400$ at \SI{33.46}{\degree}, and the cliff of
Sec.~\ref{sec:transfer}); a second capture is the cheapest repair, the MF rows of
Table~\ref{tab:main} converging $400$ of $400$ against the baseline's $370$ of $400$ in its own
four-frame mode, so the single-frame margin is a statement about the protocol as much as the
estimator. The remaining error is in the statistic: by~\eqref{eq:transport} a locally planar
cell has its pose signal absorbed by its own nuisance pair, and cells whose returns fit a plane
at $R^2>0.952$ hold a median \SI{41.5}{\percent} of the pooling weight but produce
\SI{12.4}{\percent} of the geometric term's gradient norm. No scalar re-weighting of the cells
repairs it, since no scalar family reaches the efficient per-direction weight; the next step is
a depth model whose error is not concentrated where the pose signal is.

\section{Conclusion}

The local-correlation objective of single-frame targetless calibration is rough because of
where it is evaluated, not what it measures. Evaluated as a field it can be descended, and
descending it is the largest single term of the estimator's margin over the released prior
work: $138$ of $400$ calibrations converge only under our estimator and $11$ only under the
baseline, the lead survives both reciprocal controls on the depth backbone, and one set of
constants carries its accuracy to four benchmarks. Its limit is capture range on far-field
scenes, and the remaining error is in the statistic, not the optimiser.

\section*{Acknowledgment}
Claude Opus 5 (Anthropic) was used to assist with implementing the code for the
ablation studies reported in Section~\ref{sec:ablation}, and with revising the
manuscript. The research idea and all experimental design are the authors' own.
All AI-assisted content was reviewed and verified by the authors, who take full
responsibility for the content of this article.

\bibliographystyle{IEEEtran}
\balance
\bibliography{refs}

@inproceedings{geiger2012kitti,
  author    = {Geiger, Andreas and Lenz, Philip and Urtasun, Raquel},
  title     = {Are We Ready for Autonomous Driving? {T}he {KITTI} Vision Benchmark Suite},
  booktitle = {Proc. IEEE/CVF Conf. Comput. Vis. Pattern Recognit. (CVPR)},
  pages     = {3354--3361},
  year      = {2012}
}

@inproceedings{zhang2004extrinsic,
  author    = {Zhang, Qilong and Pless, Robert},
  title     = {Extrinsic Calibration of a Camera and Laser Range Finder (Improves Camera
               Calibration)},
  booktitle = {Proc. IEEE/RSJ Int. Conf. Intell. Robots Syst. (IROS)},
  pages     = {2301--2306},
  year      = {2004}
}

@inproceedings{levinson2013automatic,
  author    = {Levinson, Jesse and Thrun, Sebastian},
  title     = {Automatic Online Calibration of Cameras and Lasers},
  booktitle = {Proc. Robot.: Sci. Syst. (RSS)},
  year      = {2013},
  note      = {doi: 10.15607/RSS.2013.IX.029}
}

@inproceedings{koide2023general,
  author    = {Koide, Kenji and Oishi, Shuji and Yokozuka, Masashi and Banno, Atsuhiko},
  title     = {General, Single-Shot, Target-Less, and Automatic {LiDAR}-Camera Extrinsic
               Calibration Toolbox},
  booktitle = {Proc. IEEE Int. Conf. Robot. Autom. (ICRA)},
  pages     = {11301--11307},
  year      = {2023}
}

@article{pandey2015automatic,
  author  = {Pandey, Gaurav and McBride, James R. and Savarese, Silvio and Eustice, Ryan M.},
  title   = {Automatic Extrinsic Calibration of Vision and {LiDAR} by Maximizing Mutual
             Information},
  journal = {J. Field Robot.},
  volume  = {32},
  number  = {5},
  pages   = {696--722},
  year    = {2015}
}

@inproceedings{borer2024chaos,
  author    = {Borer, Jack and Tschirner, Jeremy and {\"O}lsner, Florian and Milz, Stefan},
  title     = {From Chaos to Calibration: A Geometric Mutual Information Approach to
               Target-Free Camera {LiDAR} Extrinsic Calibration},
  booktitle = {Proc. IEEE/CVF Winter Conf. Appl. Comput. Vis. (WACV)},
  year      = {2024},
  note      = {arXiv:2311.01905}
}

@inproceedings{zhang2025claim,
  author    = {Zhang, Zhuo and Liu, Yonghui and Zhang, Meijie and Tan, Feiyang and Ding, Yikang},
  title     = {{CLAIM}: Camera-{LiDAR} Alignment with Intensity and Monodepth},
  booktitle = {Proc. IEEE/RSJ Int. Conf. Intell. Robots Syst. (IROS)},
  pages     = {17921--17926},
  year      = {2025},
  doi       = {10.1109/IROS60139.2025.11247484}
}

@article{huang2024mias,
  author  = {Huang, Zhiwei and Zhang, Yikang and Chen, Qijun and Fan, Rui},
  title   = {Online, Target-Free {LiDAR}-Camera Extrinsic Calibration via Cross-Modal Mask
             Matching},
  journal = {IEEE Trans. Intell. Veh.},
  volume  = {10},
  number  = {5},
  pages   = {3531--3542},
  year    = {2025}
}

@article{huang2025edo,
  author  = {Huang, Zhiwei and Li, Jiaqi and Zhao, Hongbo and Ma, Xiao and Zhong, Ping and
             Zhou, Xiao-Hu and Ye, Wei and Fan, Rui},
  title   = {Environment-Driven Online {LiDAR}-Camera Extrinsic Calibration},
  journal = {IEEE Trans. Autom. Sci. Eng.},
  volume  = {22},
  pages   = {24164--24176},
  year    = {2025},
  doi     = {10.1109/TASE.2025.3627253}
}

@inproceedings{schneider2017regnet,
  author    = {Schneider, Nick and Piewak, Florian and Stiller, Christoph and Franke, Uwe},
  title     = {{RegNet}: Multimodal Sensor Registration Using Deep Neural Networks},
  booktitle = {Proc. IEEE Intell. Veh. Symp. (IV)},
  pages     = {1803--1810},
  year      = {2017},
  note      = {arXiv:1707.03167}
}

@inproceedings{yang2024depthanythingv2,
  author  = {Yang, Lihe and others},
  title     = {Depth Anything {V2}},
  booktitle = {Proc. Adv. Neural Inf. Process. Syst. (NeurIPS)},
  year      = {2024},
  note      = {arXiv:2406.09414}
}

@inproceedings{lin2025depthanything3,
  author    = {Lin, Haotong and others},
  title     = {Depth Anything 3: Recovering the Visual Space from Any Views},
  booktitle = {Proc. Int. Conf. Learn. Represent. (ICLR)},
  year      = {2026},
  note      = {arXiv:2511.10647}
}

@inproceedings{wang2025moge2,
  author  = {Wang, Ruicheng and others},
  title     = {{MoGe-2}: Accurate Monocular Geometry with Metric Scale and Sharp Details},
  booktitle = {Proc. Adv. Neural Inf. Process. Syst. (NeurIPS)},
  year      = {2025},
  note      = {arXiv:2507.02546}
}

@book{kish1965survey,
  author    = {Kish, Leslie},
  title     = {Survey Sampling},
  publisher = {John Wiley \& Sons},
  year      = {1965}
}

@article{neyman1948consistent,
  author  = {Neyman, Jerzy and Scott, Elizabeth L.},
  title   = {Consistent Estimates Based on Partially Consistent Observations},
  journal = {Econometrica},
  volume  = {16},
  number  = {1},
  pages   = {1--32},
  year    = {1948}
}

@book{fan1996local,
  author    = {Fan, Jianqing and Gijbels, Ir{\`e}ne},
  title     = {Local Polynomial Modelling and Its Applications},
  publisher = {Chapman \& Hall},
  year      = {1996}
}

@article{studholme1999overlap,
  author  = {Studholme, Colin and Hill, Derek L. G. and Hawkes, David J.},
  title   = {An Overlap Invariant Entropy Measure of {3D} Medical Image Alignment},
  journal = {Pattern Recognit.},
  volume  = {32},
  number  = {1},
  pages   = {71--86},
  year    = {1999}
}

@inproceedings{herau2024gscalib,
  author    = {Herau, Quentin and others},
  title     = {{3DGS-Calib}: {3D} {G}aussian {S}platting for Multimodal Spatiotemporal
               Calibration},
  booktitle = {Proc. IEEE/RSJ Int. Conf. Intell. Robots Syst. (IROS)},
  pages     = {8315--8321},
  year      = {2024}
}

@article{zhou2025robust2dgs,
  author  = {Zhou, Shuyi and Xie, Shuxiang and Ishikawa, Ryoichi and Oishi, Takeshi},
  title   = {Robust {LiDAR}-Camera Calibration with {2D} {G}aussian {S}platting},
  journal = {IEEE Robot. Autom. Lett.},
  year    = {2025},
  note    = {arXiv:2504.00525}
}

@article{kim2026bridging,
  author    = {Kim, Daeho and Lee, Jeong Woon and Kwak, Kyoleen and Song, Chaneui and Hwang, Hyoseok},
  title   = {Bridging the Modality Gap with Differentiable Intensity Rendering for Online
             {LiDAR}-Camera Calibration},
  journal = {IEEE Robot. Autom. Lett.},
  year    = {2026},
  note      = {doi: 10.1109/LRA.2026.3719229}
}

@inproceedings{zhang2016degeneracy,
  title     = {On degeneracy of optimization-based state estimation problems},
  author    = {Zhang, Ji and Kaess, Michael and Singh, Sanjiv},
  booktitle = {Proc. IEEE Int. Conf. Robot. Autom. (ICRA)},
  pages     = {809--816},
  year      = {2016}
}

@article{yuan2021pixel,
  title   = {Pixel-level extrinsic self calibration of high resolution {LiDAR} and camera in targetless environments},
  author  = {Yuan, Chongjian and Liu, Xiyuan and Hong, Xiaoping and Zhang, Fu},
  journal = {IEEE Robot. Autom. Lett.},
  volume  = {6},
  number  = {4},
  pages   = {7517--7524},
  year    = {2021}
}

@article{piccinelli2025unidepthv2,
  author  = {Piccinelli, Luigi and others},
  title   = {{UniDepthV2}: Universal Monocular Metric Depth Estimation Made Simpler},
  journal = {IEEE Trans. Pattern Anal. Mach. Intell.},
  volume  = {48},
  number  = {3},
  pages   = {2354--2367},
  year    = {2026},
  doi     = {10.1109/TPAMI.2025.3628473}
}

@article{fisher1921probable,
  author  = {Fisher, Ronald A.},
  title   = {On the ``Probable Error'' of a Coefficient of Correlation Deduced from a Small
             Sample},
  journal = {Metron},
  volume  = {1},
  pages   = {3--32},
  year    = {1921}
}

@article{zhang2025higscalib,
  author  = {Zhang, Tao and Zhang, Lei and Wang, Hesheng},
  title   = {{HiGS-Calib}: A Hierarchical {3D} {G}aussian {S}platting Based Targetless
             Local-Consistent {LiDAR}-Camera Calibration Method},
  journal = {IEEE Trans. Circuits Syst. Video Technol.},
  volume  = {36},
  number  = {3},
  pages   = {3013--3026},
  year    = {2026},
  doi     = {10.1109/TCSVT.2025.3622645}
}

@inproceedings{cachier2000gaussian,
  author = {Cachier, Pascal and Pennec, Xavier},
  title = {{3D} non-rigid registration by gradient descent on a {G}aussian-windowed similarity
           measure using convolutions},
  booktitle = {Proc. IEEE Workshop Math. Methods Biomed. Image Anal. (MMBIA)},
  pages = {182--189},
  year = {2000}
}

@inproceedings{biber2003ndt,
  author = {Biber, Peter and Stra{\ss}er, Wolfgang},
  title = {The normal distributions transform: a new approach to laser scan matching},
  booktitle = {Proc. IEEE/RSJ Int. Conf. Intell. Robots Syst. (IROS)},
  pages = {2743--2748},
  year = {2003}
}

@inproceedings{koide2021vgicp,
  author = {Koide, Kenji and Yokozuka, Masashi and Oishi, Shuji and Banno, Atsuhiko},
  title = {Voxelized {GICP} for fast and accurate {3D} point cloud registration},
  booktitle = {Proc. IEEE Int. Conf. Robot. Autom. (ICRA)},
  pages = {11054--11059},
  year = {2021}
}

@article{hansen2001cma,
  author  = {Hansen, Nikolaus and Ostermeier, Andreas},
  title   = {Completely derandomized self-adaptation in evolution strategies},
  journal = {Evol. Comput.},
  volume  = {9},
  number  = {2},
  pages   = {159--195},
  year    = {2001}
}

@misc{ma2021crlf,
  author  = {Ma, Tao and Liu, Zhizheng and Yan, Guohang and Li, Yikang},
  title   = {{CRLF}: Automatic Calibration and Refinement based on Line Feature for {LiDAR} and
             Camera in Road Scenes},
  note      = {arXiv:2103.04558},
  year    = {2021}
}

@inproceedings{sun2020waymo,
  author    = {Sun, Pei and others},
  title     = {Scalability in Perception for Autonomous Driving: {W}aymo Open Dataset},
  booktitle = {Proc. IEEE/CVF Conf. Comput. Vis. Pattern Recognit. (CVPR)},
  pages     = {2443--2451},
  year      = {2020}
}

@inproceedings{lee2025zeroshotdc,
  author    = {Lee, Hyoseok and Kim, Kyeong Seon and Kwon, Byung-Ki and Oh, Tae-Hyun},
  title     = {Zero-shot Depth Completion via Test-time Alignment with Affine-invariant Depth Prior},
  booktitle = {Proc. AAAI Conf. Artif. Intell.},
  volume    = {39},
  number    = {4},
  pages     = {3877--3885},
  year      = {2025}
}

@misc{seo2026lowrankdc,
  author    = {Seo, Minseok and Lee, Wonjun and Jang, Jaehyuk and Kim, Changick},
  title     = {Efficient Test-Time Optimization for Depth Completion via Low-Rank Decoder Adaptation},
  year      = {2026},
  note      = {arXiv:2603.01765}
}

\end{document}